\documentclass[a4paper, 10pt, twocolumn]{article}

\usepackage[utf8]{inputenc}
\usepackage{titlesec}
\usepackage{amsfonts,amssymb,amsmath,amsthm}
\usepackage{subfigure}
\usepackage{graphicx}
\usepackage[footnotesize]{caption}
\usepackage{bm}
\usepackage{hyperref}
\usepackage{amsmath}
\usepackage{tikz}
\usepackage{pgfplots}
\pgfplotsset{compat=newest}
\usepackage{xcolor}
\newcommand{\norm}[1]{\left\lVert#1\right\rVert}
\DeclareMathOperator*{\minB}{min}

\titlespacing*{\section}
{0pt}{.3ex}{.3ex}
\titlespacing*{\subsection}
{0pt}{.3ex}{.3ex}

\newcommand{\argmin}[1]{\underset{#1}{\operatorname{argmin}}\;}

\DeclareMathOperator{\tr}{tr}

\graphicspath{./figures/} 

\usepackage[top=.5cm, left=1.5cm, right=1.5cm, bottom=1.5cm]{geometry}

\title{Semi-supervised dual graph regularized dictionary learning}

\author{K.-H. Tran$^1$, F.-M. Ngolè Mboula$^1$ and J.-L. Starck$^2$.\\
\footnotesize $^{1}$ CEA, LIST, Laboratoire Analyse de Donneés et l'Intélligence des Systèmes, F-91191 GIF-SUR-YVETTE Cedex, France.\\
\footnotesize $^{2}$Laboratoire AIM, UMR CEA-CNRS-Paris 7, Irfu, Service d'Astrophysique, CEA Saclay, F-91191 GIF-SUR-YVETTE Cedex, France.\
}
\date{\empty} 
\renewenvironment{abstract}{\bf\small {\em\ Abstract---}}{}
\begin{document}
\maketitle
\begin{abstract} 
In this paper, we propose a semi-supervised dictionary learning method that uses both the information in labelled and unlabelled data and jointly trains a linear classifier embedded on the sparse codes. The manifold structure of the data in the sparse code space is preserved using the same approach as the Locally Linear Embedding method (LLE). This enables one to enforce the predictive power of the unlabelled data sparse codes. We show that our approach provides significant improvements over other methods. The results can be further improved by training a simple nonlinear classifier as SVM on the sparse codes. 

\end{abstract}
\section{Introduction}
\label{sec:introduction}
\subsection{Context and related works}

\textit{Dictionary Learning} (DL) encompasses methods and algorithms that aim at deriving a set of cardinal features which enables one to concisely describe signals of a given type. The benefit of such dictionaries in sparsity-driven signal recovery has been shown in several applications (see for example \cite{elad2006image,ravishankar2011mr,beckouche2013astronomical,barthelemy2012shift}).

In numerous applications of machine learning, data are labelled and/or sampled from some regular manifold; thus it is suitable, for classification or interpolation tasks for instance, that the learned codes allow for a better discrimination of the data samples with respect to labels information or manifold's structure. The growing field of \emph{supervised dictionary learning} precisely consists of DL methods that account for these additional information (a recent review can be found in \cite{gangeh2015supervised}). Unlike the supervised classification in which only labelled data is used to train the classifier, the unlabelled data is also used in training to make use of all the manifold's structure information available. 

\subsection{Notations}
\label{sec:notations}

Given date set $\textbf{X} \in \mathbb{R}^{n\times N}$, $n$ being the number of features and $N$ the number of samples:  \textbf{X} = $[\textbf{x}_{1}, \textbf{x}_{2},...,\textbf{x}_{N}]$. We divide the data into two sets, training set made of labelled samples ($c$ classes) and testing set made of unlabelled samples: $\textbf{X} = [\textbf{X}_{train},\textbf{X}_{test}]$ which have respectively $N_{train}$ and $N_{test}$ samples. $\textbf{D} \in \mathbb{R}^{n\times p}$, the dictionary contains atoms $\textbf{d}_{i}$ $\left(i=1,...,p \right)$. We denote by $\textbf{A} = [\textbf{a}_{1}, \textbf{a}_{2},...,\textbf{a}_{N}] \in \mathbb{R}^{p\times N}$ the sparse codes matrix and by analogy we define the matrices $\textbf{A}_{train}$ and $\textbf{A}_{test}$ which contains respectively the labelled and unlabelled samples sparse codes, hence $\textbf{A} = [\textbf{A}_{train},\textbf{A}_{test}]$. $\textbf{Y}_{train} \in \mathbb{R}^{c\times N}$, matrix of known labels and $\textbf{Y}_{train}[i,j] = 1$ if $j^{th}$ sample in $i^{th}$ class, 0 otherwise. 

\section{Proposed Method}

The dictionary learning is achieved by solving a problem of the form:
\begin{equation}
\label{objective_function}
\begin{split}
&\minB_{\textbf{W},\textbf{A}, \textbf{D}\in \mathcal{C}} \norm{\textbf{X}-\textbf{D}\textbf{A} }_{F}^{2} + \lambda\norm{\textbf{A}}_{1}  +\mathcal{F}_D(\textbf{D}) + \mathcal{F}_A(\textbf{A})\\
& + \gamma\norm{\textbf{Y}_{train}-\textbf{W}\textbf{A}_{train}}_{F}^{2} + \mu\norm{\textbf{W}}_{F}^2,
\end{split}
\end{equation}
where $\mathcal{C} = \{ \textbf{D} , \norm{\textbf{d}_{i}}_2 = 1,  \forall i = 1,2,...,p \}$.

The matrix $\textbf{W} \in \mathbb{R}^{c \times p} $ is a linear classifier (LC) in the sparse code space. 
Following the idea proposed in \cite{Yankelevsky16Dec}, we want to constraint the atoms structures via the functional $\mathcal{F}_D$. On the other, we want to preserve locally the underlying samples manifold structure in the sparse code space via the functional $\mathcal{F}_A$. We detail thereafter the construction of these functionals.





\subsection{Manifold structure preservation}
\label{subsec:L_A}
To characterize the local manifold structure of the samples, we follow Locally Linear Embedding approach \cite{Roweis00}. With knn($i$) is a set contains k indices of k nearest samples from $i^{th}$ sample, we first find linear relation between each sample $\textbf{x}_i$ and its k nearest neighbours (in both training set and testing set), by solving the following problem:
\begin{equation}
\label{equa:LLE_sample}
\begin{split}
&\hat{\boldsymbol{\lambda}_{i}} = \minB_{\boldsymbol{\lambda}_{i} \in \mathbb{R}^{k} } \norm{\textbf{x}_i - \sum\limits_{j\in \text{knn}(i)} \lambda_{ij} \textbf{x}_{j}}_2^2,\\
&\text{subject to } \sum\limits_{j\in \text{knn}(i)} \lambda_{ij} = 1 \\
\end{split}
\end{equation}
Using $\hat{\boldsymbol{\lambda}_{i}}$ from Eq.~\ref{equa:LLE_sample}, we can now define $\mathcal{F}_A$ as :
\begin{equation*}
\label{equa:LLE_constraint_sparse_code}
\mathcal{F}_A(\textbf{A}) = \beta\sum\limits_{i}^N \norm{\textbf{a}_i - \sum\limits_{j\in \text{knn}(i)} \hat{\lambda}_{ij} \textbf{a}_{ij}}_2^2 
\end{equation*}
Imposing a regularity on the sparse codes distribution for both labelled and unlabelled samples is meant to enforces homogeneous label regions to be preserved in the sparse code space.

We define a matrix $\mathbf{V} \in \mathbb{R}^{N \times N}$ as $\mathbf{V}[i,j] = \hat{\lambda}_{ij}$ if  $j \in \text{knn}(i)$ and $\mathbf{V}[i,j] = 0$ otherwise.

Introducing the matrix $\mathbf{L}_A = \mathbf{I}_N-\mathbf{V}-\mathbf{V}^\top+\mathbf{V}^\top\mathbf{V}$ where $\mathbf{I}_N$ is $\mathbb{R}^{N \times N}$'s identity matrix, we can rewrite $\mathcal{F}_A(\textbf{A}) = \beta \tr(\textbf{A}\mathbf{L}_{A}\textbf{A}^\top)$. Hence the penalty $\mathcal{F}_A$ can be interpreted as a graph laplacian based regularizer as in \cite{Zheng11} or \cite{Yankelevsky16Mar}. 



\subsection{Atom structure regularization}
Following \cite{Yankelevsky16Mar}, we want to preserve features dependencies in the dictionary atoms. In other word, we want the dictionary atoms viewed as signals to be structurally similar to the samples. 

For this purpose, we define the penalty $\mathcal{F}_D$ as
$\mathcal{F}_D(\textbf{D}) =  \alpha\tr{(\textbf{D}^\top \mathbf{L}_{D} \textbf{D})}$, where the matrix $\mathbf{L}_{D}$ is obtained by solving the following problem\cite{Dong16,Yankelevsky16Mar}:
\begin{equation}
\label{2e dictionary}
\begin{split}
& \textbf{L}_{D} = \argmin {\textbf{L} \in \mathbb{R}^{n \times n}} \tr{(\textbf{X}^\top \textbf{L} \textbf{X})} + \theta \norm{\textbf{L}}_{F}^2, \\
& s.t \text{  } \textbf{L} = \textbf{L}^\top, \textbf{L}1 = 0, \tr{(\textbf{L})} = n, \textbf{L}[i,j]_{i \neq j} \leq 0.
\end{split}
\end{equation}
This amounts to learn a weighted graph that characterizes the features dependency and subsequently impose the dictionary atoms to be smooth as signals on this graph.

\section{Optimization and initialization}
\label{sec:Opti}
We use a 3 steps alternate minimization scheme:

\textbf{Sparse coding :} 
\begin{equation*}
\begin{split}
& \minB_{\textbf{A}} \norm{\textbf{X}-\textbf{D}\textbf{A} }_{F}^{2} + \beta\tr{(\textbf{A}L_{A}\textbf{A}^\top)} \\ 
&+ \gamma\norm{\textbf{Y}_{train}-\textbf{W}\textbf{A}_{train}}_{F}^{2} + \lambda\norm{\textbf{A}}_{1} \\
\end{split}
\end{equation*}

\textbf{Dictionary update :} 
\begin{equation*}
\minB_{\textbf{D} \in \mathcal{C}} \norm{\textbf{X}-\textbf{D}\textbf{A} }_{F}^{2}+\alpha\tr{(\textbf{D}^\top L_{D} \textbf{D})} \\
\end{equation*}

\textbf{Classifier update :} 
\begin{equation*}
\minB_{\textbf{W}} \gamma\norm{\textbf{Y}_{train}-\textbf{W}\textbf{A}_{train}}_{F}^{2} + \mu\norm{\textbf{W}}_{F}^2 
\end{equation*}

The first and second steps are performed using a proximal splitting method \cite{Combettes09} while the updated classifier is obtained solving first order optimality condition. The dictionary \textbf{D} is initialized randomly following a normal distribution and then each atom is normalized to have unit $l_2$ norm. The sparse codes \textbf{A} are initialized by solving a simple LASSO with dictionary \textbf{D} and $\lambda$. Finally, the LC \textbf{W} is generated by running Classifier update.    


\section{Numerical experiments}
\label{sec:Num exp}
In this section, we apply our approach Semi-Supervised Dual-Graph regularized Dictionary Learning (\textbf{SS-DG-DL}) on MNIST dataset which contains 70000 images ($28 \times 28$) of handwritten digits, from which we sample subsets of different sizes for our actual experiments. The pixels values are rescaled between 0 and 1 before processing. The hyperparameters are chosen as follows: $\gamma = 0.2, \mu = 0.1, \alpha= 10, \lambda = 0.5, \theta = 2, \beta = 0.5, k= 66 $, the number of training samples is fixed by 10000 and the number of atoms is fixed by 64 (not over-complete). To assess the benefit of using unlabelled samples in the dictionary, we run \textbf{SS-DG-DL} and and its supervised counterpart that we name \textbf{DG-DL}, which only makes use of labelled data :
\begin{equation*}
\begin{split}
&\minB_{\textbf{W},\textbf{A}_{train}, \textbf{D}\in \mathcal{C}} \norm{\textbf{X}_{train}-\textbf{D}\textbf{A}_{train}}_{F}^{2} + \alpha \tr{(\textbf{D}^\top L_{D} \textbf{D})}\\
& + \beta\tr{(\textbf{A}_{train}L_{A_{train}}\textbf{A}_{train}^\top)}  + \gamma\norm{\textbf{Y}_{train}-\textbf{W}\textbf{A}_{train}}_{F}^{2} \\
& + \mu\norm{\textbf{W}}_{F}^2 + \lambda\norm{\textbf{A}_{train}}_{1} \\
\end{split}
\end{equation*}

Unlike \textbf{SS-DG-DL} which provides directly sparse codes of testing samples after optimizing problem (\ref{objective_function}), we need to perform a sparse coding to get its codes, note that the set knn$'(i)$ contains indices of  k nearest samples in training set from $i$ testing sample: 
\begin{equation*}
\begin{split}
\label{equa:sparse_coding_DGDL}
&\minB_{\textbf{a}_{i}^{test}} \norm{\textbf{x}_{i}^{test}-\textbf{D}\textbf{a}_{i}^{test}}_{2}^{2} + \beta\norm{\textbf{a}_{i}^{test} - \hat{\lambda}_{ij}\sum\limits_{j \in \text{knn}'(i)}\textbf{a}^{train}_{j}}_2^2 \\
&+\lambda\norm{\textbf{a}_{i}^{test}}_{1}
\end{split}
\end{equation*}

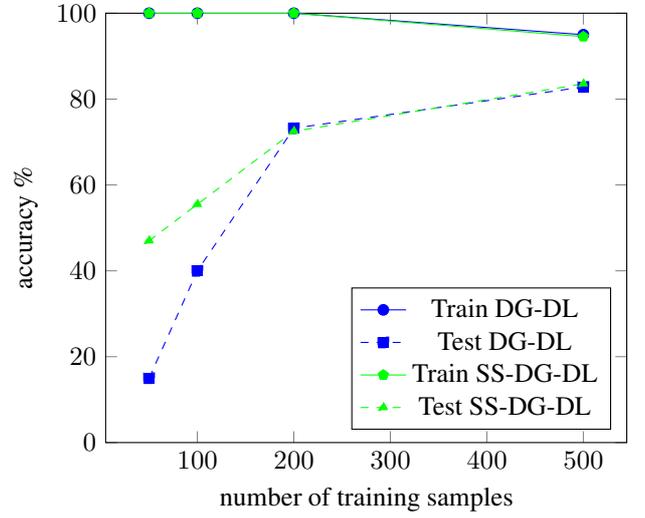
\begin{figure}
\centering
\begin{tikzpicture}
\begin{axis}
[ymax = 100, ymin = 0,
ylabel = {accuracy \%},
xlabel = {number of training samples},
xtick = {100,200,300,400,500},
legend pos = south east]
\addplot [mark = *, color = blue] coordinates {
(50,100)(100,100)(200,100)(500,95)};
\addlegendentry{Train DG-DL}
\addplot [mark = square*, color = blue,dashed] coordinates {
(50,15)(100,40)(200,73.21)(500,82.83) };
\addlegendentry{Test DG-DL}

\addplot [mark = pentagon*, color = green] coordinates {
(50,100)(100,100)(200,100)(500,94.55)};
\addlegendentry{Train SS-DG-DL}
\addplot [mark = triangle*, color = green,dashed] coordinates {
(50,47)(100,55.48)(200,72.49)(500,83.55)};
\addlegendentry{Test SS-DG-DL}

\end{axis}
\end{tikzpicture}
\caption{\textbf{DG-DL} and \textbf{SS-DG-DL} classification accuracy with various numbers of training samples evaluated by both training set and testing set.}
\label{figure_SSDGDL_DGDL_nbsamples}
\end{figure}

We test the two methods with different number of training samples (50,100,200,500), the number of unlabelled samples being fixed at 10000 and compare the classification accuracy with their respectively LC learned (\textbf{W}). In figure \ref{figure_SSDGDL_DGDL_nbsamples}, when we have a small number of training samples (50 or 100), the \textbf{SS-DG-DL}'s LC is significantly more accurate on the unlabelled samples than \textbf{DG-DL}'s. 

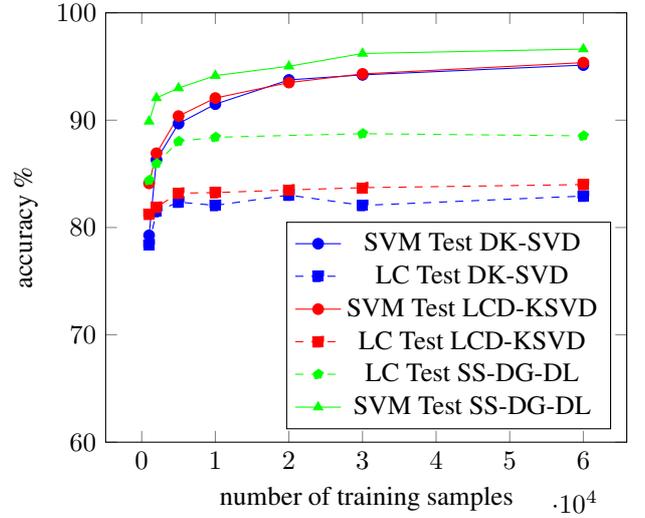
\begin{figure}
\centering
\begin{tikzpicture}
\begin{axis}
[ymax = 100, ymin = 60, xlabel = {number of training samples},
ylabel = {accuracy \%}, 
legend pos = south east]

\addplot [mark = *, color = blue] coordinates {
(1000,79.27) (2000,86.31) (5000,89.68)
(10000,91.49) (20000,93.75) (30000,94.22) (60000,95.13)};
\addlegendentry{SVM Test DK-SVD}
\addplot [mark = square*, color = blue,dashed] coordinates {
(1000,78.4) (2000,81.49) (5000,82.37)
(10000,82.06) (20000,83.02) (30000,82.06) (60000,82.93)};
\addlegendentry{LC Test DK-SVD}

\addplot [mark = *, color = red] coordinates {
(1000,84.11) (2000,86.92) (5000,90.38)
(10000,92.06) (20000,93.50) (30000,94.31) (60000,95.36)};
\addlegendentry{SVM Test LCD-KSVD}
\addplot [mark = square*, color = red,dashed] coordinates {
(1000,81.24) (2000,81.89) (5000,83.20)
(10000,83.25) (20000,83.50) (30000,83.71) (60000,84)};

\addlegendentry{LC Test LCD-KSVD}

\addplot [mark = pentagon*, color = green, dashed] coordinates {
(1000,84.4) (2000,85.95) (5000,88.02)
(10000,88.41) (30000,88.74)(60000,88.54)};
\addlegendentry{LC Test SS-DG-DL}
\addplot [mark = triangle*, color = green] coordinates {
(1000,89.89) (2000,92.07) (5000,92.99)
(10000,94.15) (20000,95.02) (30000,96.21) (60000,96.62)};
\addlegendentry{SVM Test SS-DG-DL}

\end{axis}
\end{tikzpicture}
\caption{Applying SVM to \textbf{DK-SVD}, \textbf{LCD-KSVD} and \textbf{SS-DG-DL} on sparse codes, compared to LC learned \textbf{W} in training.}
\label{figure_SVM_LC}
\end{figure}

Now we compare our approach with \textit{Discriminative K-SVD dictionary learning} (\textbf{DK-SVD}) \cite{Zhang10} and \textit{Label Consistent  K-SVD} (\textbf{LCD-KSVD}) \cite{Jiang13} with different numbers of training samples. We compare also the performance between the LC learned \textbf{W} and a simple SVM model (kernel = 'RBF') trained by using sparse codes of training samples. The result in figure~\ref{figure_SVM_LC} shows that the proposed method is $6\%$ more accurate than \textbf{DK-SVD} and \textbf{LCD-KSVD} linear classifiers-wise. As it can be seen in figure~\ref{figure_SVM_LC}, training an SVM model on the sparse codes benefits the 3 compared methods which gain between $7\%$ and $12\%$ of accuracy. However the proposed method remains the most accurate which shows that the target classes are better unfolded in its learnt sparse codes space.   
\section{Conclusion}
\label{sec:conclusion}
We introduced a semi supervised dictionary learning algorithm which, unlike current state-of-the-art supervised dictionary learning methods, makes use of both labelled and unlabelled data in jointly learning a dictionary, the sparse codes and a LC on the sparse codes space. The method was tested on the MNIST dataset. 

We show that using unlabeled data  in the training enhances the learned classifier performances, especially when few training samples are available. Moreover the proposed method significantly outperformed state-of-the-art supervised dictionary learning methods.

\section*{Acknowledgements}
This research is supported by the European Community through the grant DEDALE (contract no. 665044). 
\vspace{0.5cm }
\addcontentsline{toc}{chapter}{Bibliography}
\bibliographystyle{unsrt}
\bibliography{bib}

\end{document}